\documentclass[]{article}
\usepackage{graphicx} 
\usepackage[final]{sods_2023}
\usepackage{natbib}
\setcitestyle{authoryear} 
\usepackage[utf8]{inputenc} 
\usepackage[T1]{fontenc}    
\usepackage{hyperref}       
\usepackage{algpseudocode}
\usepackage{algorithm}
\usepackage{url}            
\usepackage{booktabs}       
\usepackage{amsfonts}       
\usepackage{nicefrac}       
\usepackage{microtype}      
\usepackage{amsmath}
\usepackage{multirow}

\usepackage{amsmath,amsfonts,bm,amssymb,mathtools,amsthm}
\usepackage{color,xcolor,xspace}
\usepackage{booktabs}
\usepackage{thm-restate}

\def\eqref#1{Eq.~(\ref{#1})}

\def\1{\bm{1}}

\def\eps{{\epsilon}}

\def\rvd{{\mathbf{d}}}

\def\rvx{{\mathbf{x}}}

\def\rmI{{\mathbf{I}}}

\def\vs{{\bm{s}}}

\def\mJ{{\bm{J}}}

\DeclareMathAlphabet{\mathsfit}{\encodingdefault}{\sfdefault}{m}{sl}
\SetMathAlphabet{\mathsfit}{bold}{\encodingdefault}{\sfdefault}{bx}{n}

\def\gD{{\mathcal{D}}}

\def\gL{{\mathcal{L}}}

\def\gN{{\mathcal{N}}}

\def\gX{{\mathcal{X}}}

\def\sR{{\mathbb{R}}}

\DeclareMathOperator*{\argmin}{arg\,min}

\begin{document}

\title{Accelerating Diffusion-based Combinatorial Optimization Solvers by Progressive Distillation}
\author{
Junwei Huang\\
Carnegie Mellon University\\
\texttt{junweih@andrew.cmu.edu}
\And
Zhiqing Sun\\
Carnegie Mellon University\\
\texttt{zhiqings@cs.cmu.edu}
\And
Yiming Yang\\
Carnegie Mellon University\\
\texttt{yiming@cs.cmu.edu}
}

\definecolor{myRed}{rgb}{1,0,0}
\definecolor{myBlue}{rgb}{0,0,1}
\definecolor{myGreen}{rgb}{0,1,0}
\definecolor{myCyan}{rgb}{0,1,1}
\newcommand{\zs}[1]{\textcolor{myBlue}{\bf\small [#1 --ZS]}}
\newcommand{\yy}[1]{\textcolor{myGreen}{\bf\small [#1 --YY]}}
\newcommand{\JH}[1]{\textcolor{myRed}{\bf\small [#1 --JH]}}

\maketitle
\begin{abstract}
   Graph-based diffusion models have shown promising results in terms of generating high-quality solutions to NP-complete (NPC) combinatorial optimization (CO) problems. However, those models are often inefficient
   in inference, due to the iterative evaluation nature of the denoising diffusion process. This paper proposes to use \textit {progressive} distillation to speed up the inference 
   by taking fewer steps (e.g., forecasting two steps ahead within a single step) during the denoising process.
   Our experimental results show that the progressively distilled model can perform inference \textbf{16} times faster with only \textbf{0.019\%} degradation in performance on the TSP-50 dataset.
\end{abstract}
\section{Introduction}
The realm of Combinatorial Optimization (CO) is intricately tied to optimization within discrete spaces. A significant portion of CO problems falls into the category of NP-complete (NPC) problems,
for which efficient algorithms for exact solutions are nearly impossible to find.
In the past, researchers have often turned to traditional approximation algorithms as a means to tackle these NPC problems~\citep{polytsp,algomis}. Unfortunately, these traditional methods demand an extensive degree of mathematical understanding specific to each individual problem, resulting in heuristics that lack transferability across diverse problem types.

Modern approaches utilize the pattern recognition power of machine learning techniques to generate high-quality solutions.
Representative methods include the deep reinforcement learning (DRL) models ~\citep{Khalil2017LearningCO,Barrett2019ExploratoryCO,Yao2021ReversibleAD,Bello2016NeuralCO} and neural diffusion models \citep{graikos2022diffusion, sun2023difusco}. DRL approaches
focus on training an agent to learn the heuristics for solving a CO problem, by allowing the agent to interact with its environment, aided by a graph representation of the problem.
In the unsupervised setting for a DRL solver, the system learns the search strategy without requiring annotated optimal solutions for training-set graphs \citep{erdos_unsupervised,Wang2022UnsupervisedLF}, but DRL CO solvers are usually hard to train, due to the sparse reward problem \citep{wu2021learning}.
In the supervised settings, on the other hand, neural network solvers learn a generative process to directly predict the high-quality solutions to CO problems \citep{joshi2019efficient,sun2023difusco}.
They are usually more stable in training and efficient in inference, but require near-optimal annotations generated by traditional solvers.

Recent progress in diffusion models for image generation~\citep{DDPM,song2020denoising,score_based} has sparked interest in applying diffusion models to CO problem with the hope of exploiting its expressiveness. \citet{graikos2022diffusion} show by encoding the traveling salesman problem (TSP) as $64 \times 64$ greyscale images, image diffusion models~\citep{DDPM} are able to generate heatmaps from which the solutions could be extracted. \citet{sun2023difusco} proposed the DIFUSCO framework, which works by explicitly modeling the graph structure of CO problems and denoising the variable indicator vector. However, these diffusion-based methods are inefficient in inference because of the non-trivial number of inference steps required to produce satisfactory results. 

To remedy the efficiency problem encountered by these diffusion-based solvers, we propose to use progressive distillation \citep{rezagholizadeh-etal-2022-pro,Salimans2022ProgressiveDF,meng2023distillation} to address the inference efficiency issue of DIFUSCO, accelerating its inference while preserving its solution quality. We train the student to compress two denoising steps from the teacher into one in a single training iteration and we iteratively perform this compression to achieve further distilled student.
We report a 0.019\% performance degradation on our 4x distilled student, which inferences 16 times faster than the teacher.

\section{Method}
\subsection{Problem Definition}
\label{tsp def}
In this paper, our primary focus is on the Travelling Salesman Problem (TSP) as a subject of investigation. However, we posit that both the algorithm we have developed and our conclusions have broader applicability, potentially extending to other types of Combinatorial Optimization (CO) problems. This perspective aligns with the findings presented in DIFUSCO \citep{sun2023difusco}.

Formally, we define the possible solution space of TSP instance \(\vs\) to be \(\gX_\vs = \{0, 1\}^N\), where \(N\) is the number of edges in the problem instance. For each possible solution vector \(\rvx \in \gX_\vs \), \(\rvx_i\) is the indicator variable for whether edge \(i\) is chosen in the solution vector \(\rvx\).
We define the objective function \(\mJ_\vs: \gX_\vs \to \sR\) such that it incorporates both the cost and the validity of the solution:
\begin{equation}
    \mJ_\vs(\rvx) = \mathbf{cost}_\vs(\rvx) + \mathbf{valid}_\vs(\rvx) 
\end{equation}
where \(\textbf{cost}_\vs(\rvx) = \rvx^T \rvd, \rvd_i\) being the \(i^{th}\) edge's weight in \(\vs\), and \(\textbf{valid}_\vs(\rvx)\) is 0 for a feasible TSP solution, \(+\infty\) for infeasible solutions. The optimal solution to \(\vs\) is 
\begin{equation}
    \rvx^* =  \argmin_{\rvx \in \gX_\vs} \mJ_\vs(\rvx)
\end{equation}
\subsection{Diffusion Solvers for Combinatorial Optimization}
DIFUSCO \citep{sun2023difusco} is a framework for directly modeling the graph representation of CO problems and solving CO problems via diffusion models. By first re-scaling~\citep{rescale} \(\{0, 1\}\) valued variables to \(\{-1,1\}\) and treating them as real-valued variables, it becomes viable to apply continuous diffusion~\citep{DDPM,song2020denoising} to the problem setup in \ref{tsp def}. Re-scaling for binary valued variable at time step \(t \in \{1, \ldots T\}\) is given as:
\begin{align}
    \hat{\rvx}_t &= 2\rvx_t - 1 
\end{align}
After the re-scaling, we then use the vanilla forward process proposed by \cite{DDPM}
\begin{align}
    q_\theta(\hat{\rvx}_t | \hat{\rvx}_{t-1}) = \mathcal{N}(\hat{\rvx}_t; \sqrt{1-\beta_t}\hat{\rvx}_{t-1}, \beta_t\rmI)
\end{align}
where \(\beta_t\) is the ratio of corruption at timestep \(t\). The \(t\) step marginal distribution of \(\hat{\rvx}_t\) is therefore:
\begin{align}
    q_\theta(\hat{\rvx}_t | \hat{\rvx}_0) = \mathcal{N}(\hat{\rvx}_t; \sqrt{\Bar{\alpha}_t} \hat{\rvx}_0, (1-\Bar{\alpha}_t )\rmI)
\end{align}
We train a neural network \(\eps_\theta(\cdot, \cdot) \) to predict the Gaussian noise \(\eps\) and given the corrupted \(\hat{\rvx}_t\) and timestep \(t\).
\begin{align}
    \Tilde{\eps_t} = \eps_\theta(\hat{\rvx}_t, t)
\end{align}
We apply the Mean Squared Error (MSE) Loss to \(\Tilde{\eps}_t\) and \(\eps_t\) and optimize \(\eps_\theta\) with respect to this loss, producing a performant predictor of noise that we could use to iteratively denoise and produce the predicted solution \(\hat{\rvx}_0\). The predicted solution \(\hat{\rvx}_0\) is obtained with the Gaussian Posterior from Denoising Diffusion Implicit Model (DDIM)~\citep{song2020denoising}.
\begin{align}
    \text{GuassianPosterior}(\Tilde{\eps}_t, \hat{\rvx}_t) = \sqrt{\frac{\Bar{\alpha}_{t-1}}{\Bar{\alpha}_{t}}}\left(\hat{\rvx}_t - \sqrt{1 - \Bar{\alpha}_t}\Tilde{\eps}_t\right) + \sqrt{1- \Bar{\alpha}_{t-1}}\Tilde{\eps}_t
\end{align}

We finally perform quantization to obtain \(\rvx_0\) from \(\hat{\rvx}_0\).

\subsection{Progressive Distillation}
The DIFUSCO framework is capable of producing high-quality solutions for CO problems like TSP-50, TSP-100, etc. However, this framework is often inefficient in inference due to the need of a considerable number of inference diffusion steps to generate reasonable solutions. Aiming to accelerate this method while preserving solution quality, we propose to use progressive distillation to accelerate its inference.

Progressive Distillation~\citep{rezagholizadeh-etal-2022-pro,Salimans2022ProgressiveDF} is an algorithm designed to accelerate the inference speed of diffusion models. For one iteration of progressive distillation, we update the student model to make its one denoising step match two denoising steps from the teacher model. We then iteratively train a series of student models in this manner, where the teacher model in each iteration is initialized from the converged student model in the previous iteration. The objective is to progressively distill knowledge from one model to the next, gradually improving the performance of the student model with fewer inference steps.

By the end of multiple distillation iterations, it becomes viable for the yielded progressively distilled student model to compress the diffusion steps required to achieve the same level of quality of solution by multiple times. We describe the progressive distillation training algorithm for DIFUSCO in \ref{alg:cap}.
\begin{algorithm}[t]
\footnotesize
\caption{Progressive Distillation on DIFUSCO - training}\label{alg:cap}
\begin{algorithmic}
\Require Trained teacher model $\widehat{\eps_\eta}(\cdot,\cdot)$
\Require Data Set \( \gD\), Student sampling steps $N$, Distillation factor $K$
\For{$K$ iterations}
        \State $\theta \gets \eta$ \Comment{Init student from teacher}
        \While{not converged}
            \State \textbf{Sample} \{ $\rvx \sim \gD$, \ \ \  \(t = i/N,\ \ \  i \sim \) Categorical \([1,2,\ldots,N]\),  \ \ \ \(\epsilon \sim \gN(0,\rmI)\) \}
            \State \(\rvx_t = \alpha_t \rvx + \sigma_t \epsilon\)
            \State \(t' = t - 0.5/N, t'' = t - 1/N\)
            \State \textbf{\#2 steps of teacher denoising}
            \State \(\rvx_{t'} = \)\ GaussianPosterior$_{\eta}\left[\hat{\eps_\eta}(\rvx_t, t),\rvx_t\right]$
            \State \(\rvx_{t''} = \)\ GaussianPosterior$_{\eta}\left[\hat{\epsilon_\eta}(\rvx_{t'}, {t'}),\rvx_{t'}\right]$
            \State \textbf{\#1 step of student denoising}
            \State \(\Tilde{\rvx} = \) \ GaussianPosterior\(_{\theta} \left[\hat{\epsilon_\theta}(\rvx_t, t), \rvx_t\right] \)
            \State $\gL =  ||\rvx_{t''} - \Tilde{\rvx}||^2_2$  \Comment{MSE loss}
            \State $\theta \gets \theta - \gamma \nabla_\theta \gL$ \Comment{Gradient Update on student model parameters}
        \EndWhile
\State $\eta \gets \theta$ \Comment{Student becomes the new teacher}
\State $N \gets N/2$ \Comment{ Halve number of sampling steps}
\EndFor
\end{algorithmic}
\end{algorithm}

\section{Experiments}
We benchmark our results against the TSP-50 dataset, where each training sample \(\rvx_0\) is the solution to a TSP instance with exactly 50 vertices in the problem graph. We fixed the diffusion steps to 1024 in all experiments. For the teacher model and each student model, we vary the number of inference steps as powers of 2, from 4 to 64 steps. The result of the distilled students after $1\times$, $2\times$, $3\times$, and $4\times$ progressive distillation iterations are shown in Figure 1. We also perform \(4\times\) parallel sampling for all experiments in Figure 1 and report results in Figure 2. $1\times$, $2\times$, $3\times$, and $4\times$ distilled students effectively have 512, 256, 128, and 64 diffusion steps during training.

\paragraph{Model Setup} Following \citet{sun2023difusco}, we set the noise predictor neural network \(\hat{\eps}_\theta(\cdot,\cdot)\) to be a 12-layer anisotropic Graph Neural Network~\citep{bresson2018experimental} with a width of 256. We fix the diffusion scheduler \(\{\beta_i\}_{i=1}^{T}\) to be a linear denoising schedule from \(\beta_1 = 10^{-4} \) to \(\beta_T = 0.02\). We use DDIM~\cite{song2020denoising} for fast inference and use a linear skipping strategy for all experiments \citep{sun2023difusco}.

\paragraph{Performance Metrics}Per convention, we use the solution cost drop percentage on the test set as the metric of performance. Solution cost drop percentage is calculated as the the difference between yielded solution cost and ground truth cost divided by the ground truth cost.


\paragraph{Same inference steps}
We first examine how much better our distilled student performs when the teacher and the student are allowed the same number of inference steps.
As shown in Figure 1, we observed that our distilled student performs considerably better than the teacher when we allow the same number of inference steps for the student and teacher. The student is expected to compress multiple steps of the teacher into one step of its own. Therefore the student inferencing with a distill factor of \(k\) is theoretically equivalent to teacher inferencing with \(k\) times the student inference steps. When parallel sampling is enabled, we observe that performance gap between teacher and distilled students shrinks but does not vanish.

\paragraph{Inference with \(1/k\) of steps}
We also examine how minimal the performance degradation is when we restrict the student to inference with \(1/k\) of teacher inference steps
We observed that distilled students are able to retain minimal performance degradation compared to our already optimized teacher. Our 4x distilled student was able to compress 64 inference steps of teacher model into 4 steps. Comparing the result of student inferencing with 4 steps and teacher inferencing with 64 steps, we observed a performance degradation of 0.019\%.

\section{Related Work}
 Prior work by~\cite{graikos2022diffusion} used image diffusion model to introduce visual inductive bias to aid solving CO problems. 
We would like to point out progressive distillation on image generation models has been explored by~\cite{Salimans2022ProgressiveDF}. We explored upon the prediction target mentioned and decided to adopt unscaled clean solution as prediction target due to its stability during experimentation.

\begin{figure}
    \centering
    \begin{minipage}{0.47\textwidth}
        \includegraphics[width=1\textwidth]{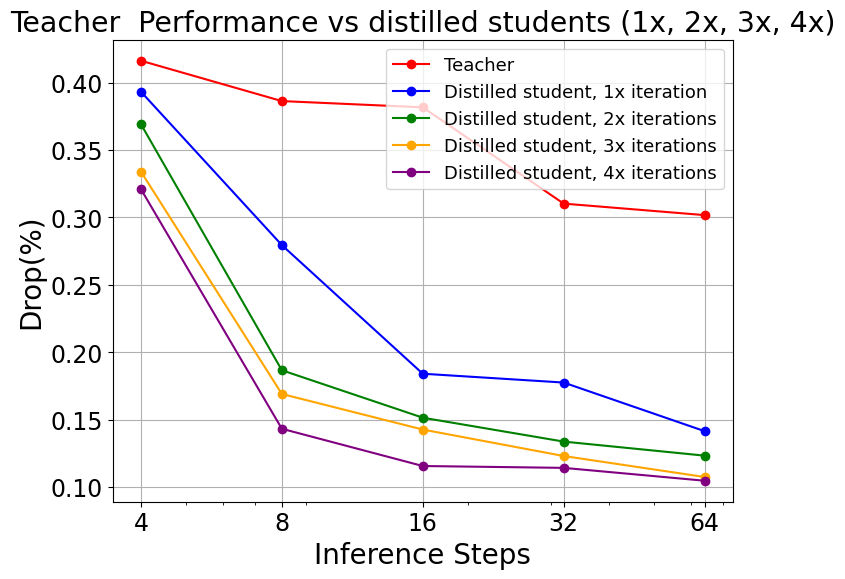}
        \hspace{-0.03\textwidth}
    \label{fig:regular}
    \centering
    \end{minipage}
    \hfill
    \begin{minipage}{0.47\textwidth}
    \includegraphics[width=1\textwidth]{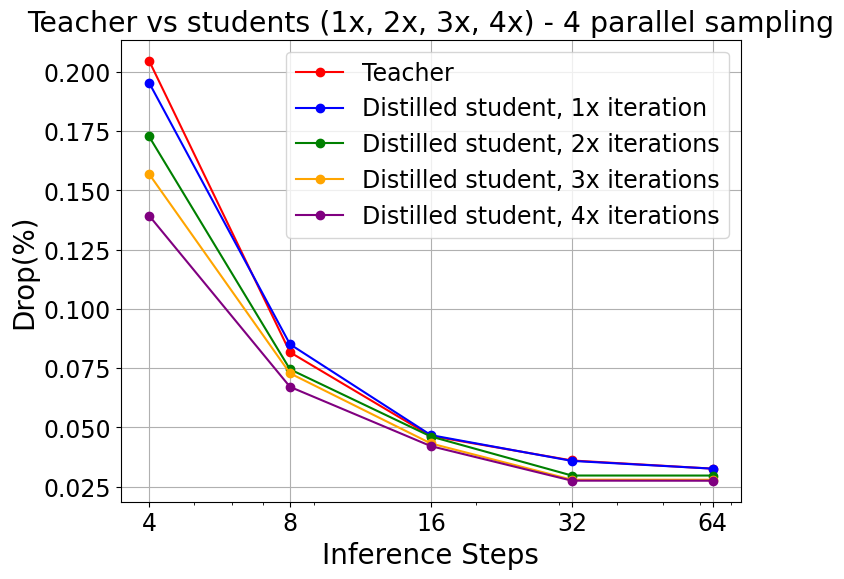}
    \end{minipage}
    \caption{TSP-50 results for $1\times$, $2\times$, $3\times$, and $4\times$ distilled student model performance against an optimized teacher model performance. We vary the number of inference steps and tabulate performance drop on each choice of inference step. Figure 2: We enable $4\times$ parallel sampling where we decode 4 samples in parallel and pick the best candidate.}
    \label{fig:parallel}
\end{figure}

\section{Discussion \& Conclusion}
In this paper, we have demonstrated that the adoption of a progressive distillation scheme can significantly accelerate the inference process of diffusion-based CO solvers. Crucially, this enhancement in speed does not compromise the quality of the solutions obtained, resulting in only a negligible performance degradation. While our current experimentation is focused on TSP-50, we are optimistic about the potential of extending our methodology to other CO problems in future studies. This area of research provides a promising direction for future exploration and analysis.

Aside from continuous Gaussian noises, DIFUSCO \citep{sun2023difusco} also mentioned injecting discrete Bernoulli noise which yielded better experimental results. We will extend our distillation scheme to support Bernoulli noise in future work, taking advantage of its inherent consistency with the discrete nature of CO problems. We also consider exploring other performant backbone networks, such as transformers~\citep{Peebles2022ScalableDM}.

\bibliography{sods}

\begin{thebibliography}{21}
\providecommand{\natexlab}[1]{#1}
\providecommand{\url}[1]{\texttt{#1}}
\expandafter\ifx\csname urlstyle\endcsname\relax
  \providecommand{\doi}[1]{doi: #1}\else
  \providecommand{\doi}{doi: \begingroup \urlstyle{rm}\Url}\fi

\bibitem[Arora(1998)]{polytsp}
S.~Arora.
\newblock Polynomial time approximation schemes for euclidean traveling
  salesman and other geometric problems.
\newblock \emph{J. ACM}, 45\penalty0 (5):\penalty0 753–782, sep 1998.
\newblock ISSN 0004-5411.
\newblock \doi{10.1145/290179.290180}.
\newblock URL \url{https://doi.org/10.1145/290179.290180}.

\bibitem[Barrett et~al.(2019)Barrett, Clements, Foerster, and
  Lvovsky]{Barrett2019ExploratoryCO}
T.~D. Barrett, W.~R. Clements, J.~N. Foerster, and A.~I. Lvovsky.
\newblock Exploratory combinatorial optimization with reinforcement learning.
\newblock \emph{ArXiv}, abs/1909.04063, 2019.

\bibitem[Bello et~al.(2016)Bello, Pham, Le, Norouzi, and
  Bengio]{Bello2016NeuralCO}
I.~Bello, H.~Pham, Q.~V. Le, M.~Norouzi, and S.~Bengio.
\newblock Neural combinatorial optimization with reinforcement learning.
\newblock \emph{ArXiv}, abs/1611.09940, 2016.

\bibitem[Bresson and Laurent(2018)]{bresson2018experimental}
X.~Bresson and T.~Laurent.
\newblock An experimental study of neural networks for variable graphs.
\newblock In \emph{International Conference on Learning Representations
  Workshop}, 2018.

\bibitem[Chen and Tian(2019)]{rescale}
X.~Chen and Y.~Tian.
\newblock Learning to perform local rewriting for combinatorial optimization.
\newblock In H.~Wallach, H.~Larochelle, A.~Beygelzimer, F.~d\textquotesingle
  Alch\'{e}-Buc, E.~Fox, and R.~Garnett, editors, \emph{Advances in Neural
  Information Processing Systems}, volume~32. Curran Associates, Inc., 2019.
\newblock URL
  \url{https://proceedings.neurips.cc/paper_files/paper/2019/file/131f383b434fdf48079bff1e44e2d9a5-Paper.pdf}.

\bibitem[Graikos et~al.(2022)Graikos, Malkin, Jojic, and
  Samaras]{graikos2022diffusion}
A.~Graikos, N.~Malkin, N.~Jojic, and D.~Samaras.
\newblock Diffusion models as plug-and-play priors.
\newblock In \emph{Thirty-Sixth Conference on Neural Information Processing
  Systems}, 2022.
\newblock URL \url{https://arxiv.org/pdf/2206.09012.pdf}.

\bibitem[Ho et~al.(2020)Ho, Jain, and Abbeel]{DDPM}
J.~Ho, A.~Jain, and P.~Abbeel.
\newblock Denoising diffusion probabilistic models.
\newblock \emph{ArXiv}, abs/2006.11239, 2020.

\bibitem[Joshi et~al.(2019)Joshi, Laurent, and Bresson]{joshi2019efficient}
C.~K. Joshi, T.~Laurent, and X.~Bresson.
\newblock An efficient graph convolutional network technique for the travelling
  salesman problem.
\newblock \emph{arXiv preprint arXiv:1906.01227}, 2019.

\bibitem[Karalias and Loukas(2020)]{erdos_unsupervised}
N.~Karalias and A.~Loukas.
\newblock Erdos goes neural: an unsupervised learning framework for
  combinatorial optimization on graphs.
\newblock In H.~Larochelle, M.~Ranzato, R.~Hadsell, M.~Balcan, and H.~Lin,
  editors, \emph{Advances in Neural Information Processing Systems}, volume~33,
  pages 6659--6672. Curran Associates, Inc., 2020.
\newblock URL
  \url{https://proceedings.neurips.cc/paper_files/paper/2020/file/49f85a9ed090b20c8bed85a5923c669f-Paper.pdf}.

\bibitem[Khalil et~al.(2017)Khalil, Dai, Zhang, Dilkina, and
  Song]{Khalil2017LearningCO}
E.~B. Khalil, H.~Dai, Y.~Zhang, B.~N. Dilkina, and L.~Song.
\newblock Learning combinatorial optimization algorithms over graphs.
\newblock In \emph{NIPS}, 2017.

\bibitem[Lawler et~al.(1980)Lawler, Lenstra, and Rinnooy~Kan]{algomis}
E.~L. Lawler, J.~K. Lenstra, and A.~H.~G. Rinnooy~Kan.
\newblock Generating all maximal independent sets: Np-hardness and
  polynomial-time algorithms.
\newblock \emph{SIAM Journal on Computing}, 9\penalty0 (3):\penalty0 558--565,
  1980.
\newblock \doi{10.1137/0209042}.
\newblock URL \url{https://doi.org/10.1137/0209042}.

\bibitem[Meng et~al.(2023)Meng, Rombach, Gao, Kingma, Ermon, Ho, and
  Salimans]{meng2023distillation}
C.~Meng, R.~Rombach, R.~Gao, D.~Kingma, S.~Ermon, J.~Ho, and T.~Salimans.
\newblock On distillation of guided diffusion models.
\newblock In \emph{Proceedings of the IEEE/CVF Conference on Computer Vision
  and Pattern Recognition}, pages 14297--14306, 2023.

\bibitem[Peebles and Xie(2022)]{Peebles2022ScalableDM}
W.~S. Peebles and S.~Xie.
\newblock Scalable diffusion models with transformers.
\newblock \emph{ArXiv}, abs/2212.09748, 2022.

\bibitem[Rezagholizadeh et~al.(2022)Rezagholizadeh, Jafari, Saladi, Sharma,
  Pasand, and Ghodsi]{rezagholizadeh-etal-2022-pro}
M.~Rezagholizadeh, A.~Jafari, P.~S. Saladi, P.~Sharma, A.~S. Pasand, and
  A.~Ghodsi.
\newblock Pro-{KD}: Progressive distillation by following the footsteps of the
  teacher.
\newblock In \emph{Proceedings of the 29th International Conference on
  Computational Linguistics}, pages 4714--4727, Gyeongju, Republic of Korea,
  Oct. 2022. International Committee on Computational Linguistics.
\newblock URL \url{https://aclanthology.org/2022.coling-1.418}.

\bibitem[Salimans and Ho(2022)]{Salimans2022ProgressiveDF}
T.~Salimans and J.~Ho.
\newblock Progressive distillation for fast sampling of diffusion models.
\newblock \emph{ArXiv}, abs/2202.00512, 2022.

\bibitem[Song et~al.(2020)Song, Meng, and Ermon]{song2020denoising}
J.~Song, C.~Meng, and S.~Ermon.
\newblock Denoising diffusion implicit models.
\newblock \emph{arXiv:2010.02502}, October 2020.
\newblock URL \url{https://arxiv.org/abs/2010.02502}.

\bibitem[Song and Ermon(2020)]{score_based}
Y.~Song and S.~Ermon.
\newblock Improved techniques for training score-based generative models.
\newblock In H.~Larochelle, M.~Ranzato, R.~Hadsell, M.~Balcan, and H.~Lin,
  editors, \emph{Advances in Neural Information Processing Systems}, volume~33,
  pages 12438--12448. Curran Associates, Inc., 2020.
\newblock URL
  \url{https://proceedings.neurips.cc/paper_files/paper/2020/file/92c3b916311a5517d9290576e3ea37ad-Paper.pdf}.

\bibitem[Sun and Yang(2023)]{sun2023difusco}
Z.~Sun and Y.~Yang.
\newblock Difusco: Graph-based diffusion solvers for combinatorial
  optimization, 2023.

\bibitem[Wang et~al.(2022)Wang, Wu, Yang, Hao, and Li]{Wang2022UnsupervisedLF}
H.~Wang, N.~Wu, H.~Yang, C.~Hao, and P.~Li.
\newblock Unsupervised learning for combinatorial optimization with principled
  objective relaxation.
\newblock \emph{ArXiv}, abs/2207.05984, 2022.

\bibitem[Wu et~al.(2021)Wu, Song, Cao, Zhang, and Lim]{wu2021learning}
Y.~Wu, W.~Song, Z.~Cao, J.~Zhang, and A.~Lim.
\newblock Learning improvement heuristics for solving routing problems.
\newblock \emph{IEEE transactions on neural networks and learning systems},
  33\penalty0 (9):\penalty0 5057--5069, 2021.

\bibitem[Yao et~al.(2021)Yao, Cai, and Wang]{Yao2021ReversibleAD}
F.~Yao, R.~Cai, and H.~Wang.
\newblock Reversible action design for combinatorial optimization with
  reinforcement learning.
\newblock \emph{ArXiv}, abs/2102.07210, 2021.

\end{thebibliography}

\end{document}